\documentclass{article} 
\usepackage{iclr2025_conference,times}

\usepackage{amsmath,amsfonts,bm}

\def\eqref#1{equation~\ref{#1}}

\def\1{\bm{1}}

\DeclareMathAlphabet{\mathsfit}{\encodingdefault}{\sfdefault}{m}{sl}
\SetMathAlphabet{\mathsfit}{bold}{\encodingdefault}{\sfdefault}{bx}{n}

\usepackage{hyperref}
\usepackage{url}
\usepackage{soul}
\usepackage{cleveref}
\usepackage{pifont}
\usepackage{verbatim}
\usepackage{booktabs}
\usepackage{threeparttable}
\usepackage{graphicx}
\usepackage{wrapfig}
\usepackage{listings}
\usepackage{enumitem}
\setlist[itemize]{leftmargin=*}
\setlist[enumerate]{leftmargin=*}

\newcommand{\cmark}{\ding{51}}
\newcommand{\xmark}{\ding{55}}

\title{Typhoon T1: An Open Thai Reasoning Model}

\author{Pittawat Taveekitworachai, Potsawee Manakul,\\
\textbf{Kasima Tharnpipitchai, and Kunat Pipatanakul}\\
SCB 10X R\&D\\
SCBX Group\\
Bangkok, Thailand \\
\texttt{\{pittawat,potsawee,kasima,kunat\}@scb10x.com}
}

\iclrfinalcopy

\begin{document}

\maketitle

\begin{abstract}
This paper introduces Typhoon T1, an \textit{open effort} to develop an open Thai reasoning model. A reasoning model is a relatively new type of generative model built on top of large language models (LLMs). A reasoning model generates a long chain of thought before arriving at a final answer, an approach found to improve performance on complex tasks. However, details on developing such a model are limited, especially for reasoning models that can generate traces in a low-resource language.
Typhoon T1 presents an open effort that dives into the details of developing a reasoning model in a more cost-effective way by leveraging \textit{supervised fine-tuning} using open datasets, instead of reinforcement learning. This paper shares the details about synthetic data generation and training, as well as our dataset and model weights. Additionally, we provide insights gained from developing a reasoning model that generalizes across domains and is capable of generating reasoning traces in a low-resource language, using Thai as an example. We hope this open effort provides a foundation for further research in this field.\footnote{Research artifacts: \url{https://huggingface.co/collections/scb10x/typhoon-t1-iclr-2025-sci-fm-artifacts-67e399bba281c900d778cc3e}}
\end{abstract}

\section{Introduction}\label{sec:intro}

Recent advancements in the field of large language models (LLMs) have gained more attention from the paradigm of scaling at test time \citep{snell2024scalingllmtesttimecompute,deepseekai2025deepseekr1incentivizingreasoningcapability}, which improves the performance of an LLM by allocating compute resources during test time--i.e., generating more tokens--to provide higher compute. While there exist techniques like chain-of-thought (CoT) prompting \citep{wei2022chain}, self-consistency \citep{wang2023selfconsistency}, and best-of-N evaluation \citep{snell2024scalingllmtesttimecompute}, which do not change the weights of the LM and focus more on prompting and sampling strategies, another approach is the \textbf{reasoning model}. A reasoning model, also known as a thinking LLM \citep{wu2024thinkingllmsgeneralinstruction}, is a relatively new type of LLM that is able to generate a long reasoning trace before providing a conclusive answer, leading to improved performance across tasks.

The long reasoning trace of a reasoning model often consists of behaviors such as breaking down a task into sub-tasks, reflecting on its own intermediate results, and self-correcting them \citep{deepseekai2025deepseekr1incentivizingreasoningcapability}. Models that fall into this category include OpenAI's o-series\footnote{\url{https://openai.com/index/learning-to-reason-with-llms/}}, Qwen's QwQ\footnote{\url{https://qwenlm.github.io/blog/qwq-32b-preview/}}, and DeepSeek R1 \citep{deepseekai2025deepseekr1incentivizingreasoningcapability}. Although some of these studies and/or models \citep{wu2024thinkingllmsgeneralinstruction,muennighoff2025s1simpletesttimescaling,guan2025rstarmathsmallllmsmaster} are open to a certain degree, most are limited to only open-weight availability and obscure details, making them difficult to replicate in an open environment and to further progress research effort.

In this paper, we provide a detailed report on our lessons learned from developing a Thai reasoning model. We aim to develop a model capable of reasoning across tasks and generating reasoning traces in Thai, a low-resource language. In addition to the insights provided in this paper, we also open our datasets, data pipeline, training configurations, and model weights to support future research.

We name our reasoning model \textbf{Typhoon T1} to honor the open Thai LLM we selected as our initial model, \textbf{Typhoon 2 3B Instruct} \citep{pipatanakul2024typhoon2familyopen}, for developing the Thai reasoning model. We utilize a supervised fine-tuning (SFT) approach using the constructed long-thought data, presenting an alternative to large-scale reinforcement learning (RL), which was previously introduced in DeepSeek R1 \citep{deepseekai2025deepseekr1incentivizingreasoningcapability} and Skywork o1 \citep{skyworkopeno12024} as methods capable of eliciting long reasoning behaviors.

RL is often unstable \citep{NEURIPS2023_a85b405e}, which is why we selected the SFT approach for experimentation in this paper. Given the relatively new field of reasoning models, we believe there is more than one approach to achieving a reasoning model, e.g., SFT \citep{muennighoff2025s1simpletesttimescaling,qin2024o1replicationjourneystrategic}, RL \citep{deepseekai2025deepseekr1incentivizingreasoningcapability,skyworkopeno12024}, and knowledge distillation \citep{skyt12025,bespokestratos,huang2024o1replicationjourney}. We also experiment with different reasoning (i.e., thinking) formats to investigate whether introducing additional auxiliary structural XML tags to the thoughts helps improve reasoning performance. \Cref{tab:comparison} shows a comparison of openness between popular reasoning models available on the market.

\begin{table}[tbp]
    \centering
    \small
    \begin{threeparttable}
        \caption{A comparison of openness among popular reasoning models, focusing on dataset availability, data processing transparency, training methodology, and model accessibility. \textbf{P} denotes partial details. Typhoon T1 is the only model providing \textit{full openness} across all categories, including its data recipe.}
        \label{tab:comparison}
        \begin{tabular}{lcccc}
            \hline
            \bf Model & 
            \multicolumn{1}{c}{\bf Datasets} & 
            \multicolumn{1}{c}{\bf Data Recipe} & 
            \multicolumn{1}{c}{\bf Training Recipe} & 
            \multicolumn{1}{c}{\bf Model Weights} \\
            \hline
            OpenAI's o-series & \xmark & \xmark & \xmark & \xmark \\
            Google's Gemini 2.0 Flash Thinking & \xmark & \xmark & \xmark & \xmark \\
            Qwen's QwQ & \xmark & \xmark & \xmark & \cmark \\
            DeepSeek R1 & \xmark & \xmark & P & \cmark \\
            \hline
            \bf Typhoon T1 & \cmark & \cmark & \cmark & \cmark \\
            \hline
        \end{tabular}
    \end{threeparttable}
\end{table}

Our contributions are as follows:
\begin{itemize}
    \item We totally open all aspects of our approach in developing Typhoon T1, a Thai reasoning model.
    \item We conduct a comprehensive ablation study to evaluate impact of thinking formats, dataset sizes, data mixture, and language on Typhoon T1 and its variants.
\end{itemize}

\section{Methodology}\label{sec:methods}

\begin{figure}[tbp]
\begin{center}
\includegraphics[width=\linewidth]{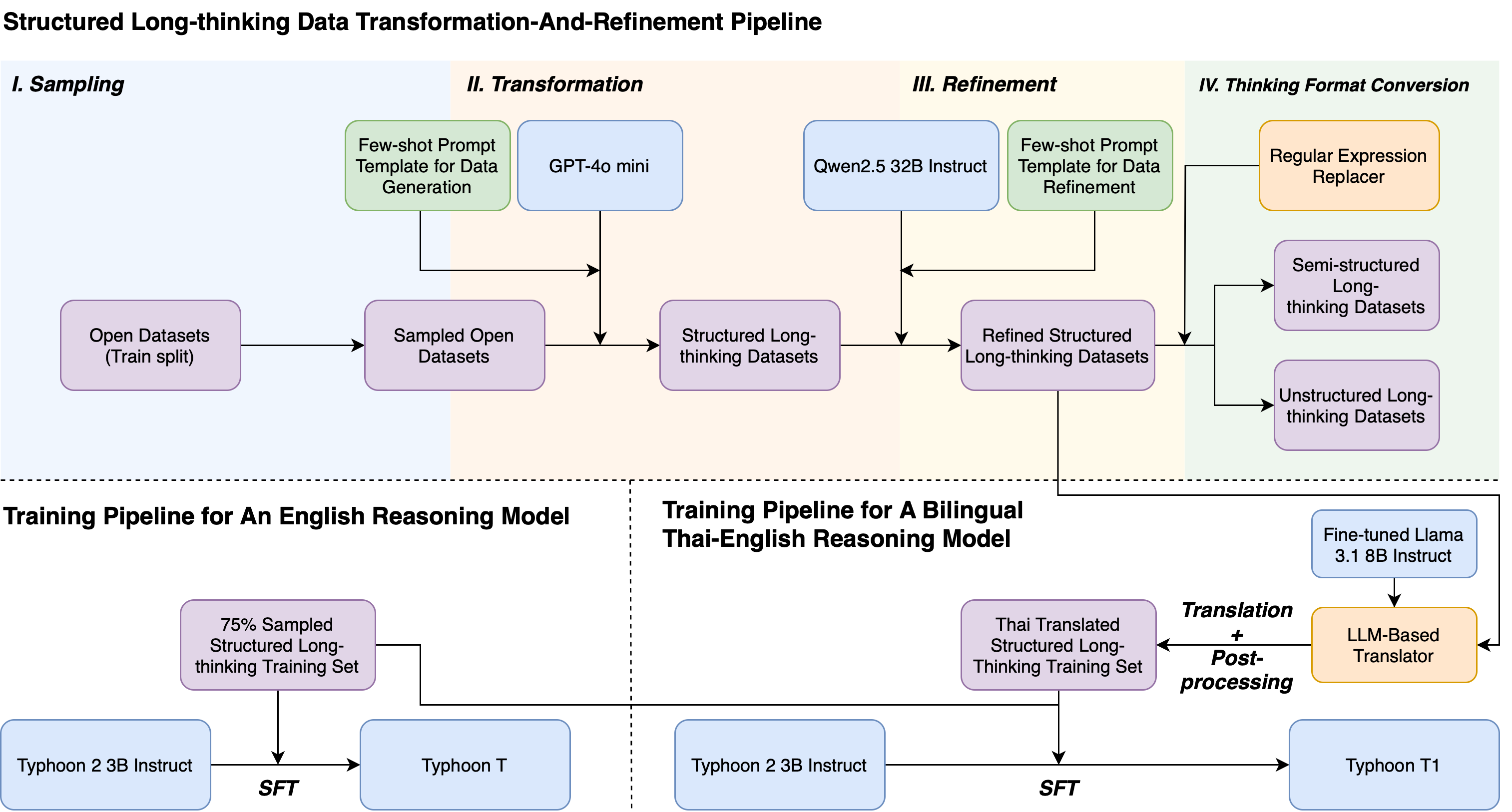}
\end{center}
\setlength{\belowcaptionskip}{-10pt}
\caption{\textit{Top:} The transformation-and-refinement pipeline used for long-thinking data generation described in Sections~\ref{sec:data_mixture} and \ref{sec:data_pipeline}. \textit{Bottom-Left:} The structured long-thinking (the best thinking format) training pipeline for Typhoon T, as described in \Cref{sec:thinking_formats}. \textit{Bottom-Right:} The bilingual English-Thai Typhoon T1 model training pipeline detailed in \Cref{sec:thai_thinking}.}
\label{fig:overview}
\end{figure}

An overview of our methodology for data generation and experiments is illustrated in \Cref{fig:overview}. In the remainder of this section, we explain the rationale for our base model selection and details of structured thinking, a new thinking format we introduce in this paper. Additionally, we provide details on the selected benchmarks used for evaluation across our experiments.

To develop a reasoning model, we first select an initial LLM. To enhance Thai performance, we consider various open-weight Thai LLMs, including OpenThaiGPT 1.5 \citep{yuenyong2024openthaigpt15thaicentricopen}, Pathumma 1.0.0 \footnote{\url{https://huggingface.co/nectec/Pathumma-llm-text-1.0.0}}, and Typhoon 2 \citep{pipatanakul2024typhoon2familyopen}. Due to resource limitations, we choose \textbf{Typhoon 2 3B Instruct} as our baseline, as it is the only available open-weight 3B model. Typhoon 2 3B is based on Llama 3.2 3B Instruct \citep{grattafiori2024llama3herdmodels}, with additional training to improve its Thai performance \citep{pipatanakul2024typhoon2familyopen}.

We select the instruct variant of the model, unlike DeepSeek R1 Zero \citep{deepseekai2025deepseekr1incentivizingreasoningcapability}, which is built on DeepSeek V3 Base, as we do not utilize RL. Our rationale is to start with a model that already follows instructions and enhance its ability to produce correct answers with long reasoning traces. Choosing an instruct model is more effective in this case than a base model, which would require a larger dataset to teach basic instruction-following skills.

To SFT the LLM into a reasoning model, we begin by preparing our training dataset. Instead of performing knowledge distillation from other reasoning models \citep{skyt12025,bespokestratos}, we opt to construct long reasoning traces ourselves. We develop a transformation-and-refinement pipeline using few-shot prompting \citep{NEURIPS2020_1457c0d6} to transform and refine a normal response into a long-form response. This allows better control over thought quality across tasks and avoids dependence on the constraints of a teacher model.

We select open datasets and refine the provided answers into long thought formats using an LLM-based pipeline. Once we obtain long-thought data, we apply SFT to the model. However, one key question remains: \textit{what is the most effective thinking format?}

We identify two main reasoning formats dominating the field: (1) unstructured thinking and (2) semi-structured thinking. In unstructured thinking, the model generates a long response without explicit separation between the chain of thought and the answer. Examples include Qwen's QwQ and Skywork o1 \citep{skyworkopeno12024}. Most existing reasoning models \citep{bespokestratos,skyt12025,deepseekai2025deepseekr1incentivizingreasoningcapability,wu2024thinkingllmsgeneralinstruction} fall into the semi-structured thinking category, where explicit separators, e.g., \texttt{<think></think>}, \texttt{<answer></answer>}, and \texttt{<R>}, distinguish the thinking process from the final answer.

\begin{figure}[tbp]
\begin{center}
\includegraphics[width=\linewidth]{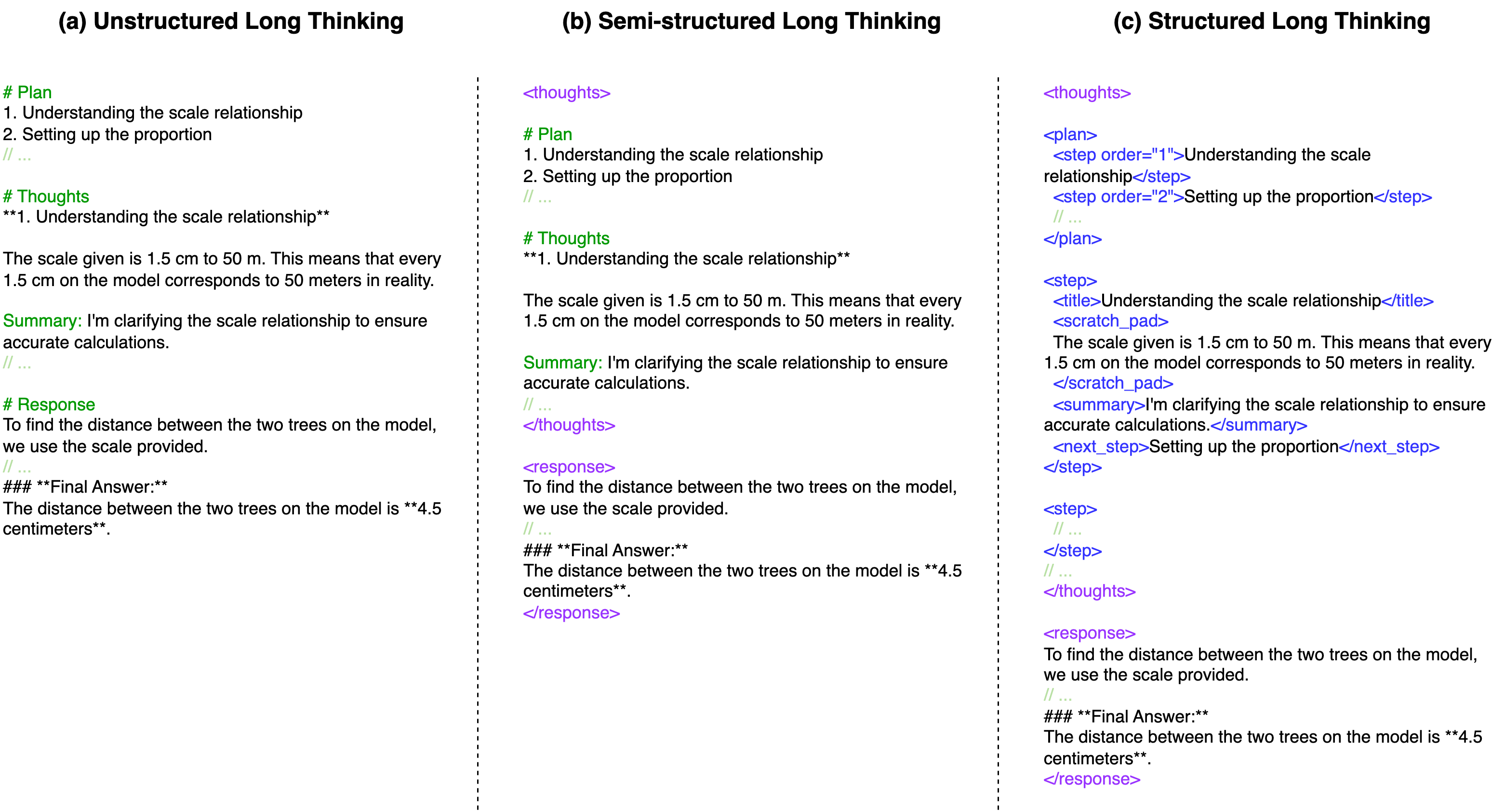}
\end{center}
\setlength{\belowcaptionskip}{-10pt}
\caption{Differences between three thinking formats: (a) Unstructured thinking, where no XML structural tags are included; (b) Semi-structured thinking, which is similar to unstructured thinking but adds \texttt{<thoughts>} and \texttt{<response>} tags to separate thoughts and responses; (c) Structured thinking, which introduces additional XML tags for structural purposes in the thoughts section.}
\label{fig:thinking_format}
\end{figure}

We observe some prompt engineering approaches \citep{nye2021workscratchpadsintermediatecomputation,ziqi-lu-2023-tab} using auxiliary tokens such as XML tags or Markdown to guide model responses. Inspired by these techniques, we introduce \textbf{structured thinking}, a new thinking format where reasoning models use auxiliary tags to generate plans and reasoning traces in each step through a scratchpad. Differences between three thinking formats is illustrated in \ref{fig:thinking_format}.

The following subsection provides further details on structured thinking. Our research question is: \textit{Among the three thinking formats, which best improves model performance?} To reduce variables in our analysis, we begin with an English-only dataset.

\subsection{Structured Thinking}\label{sec:structured_think}
Structured thinking is an approach mostly inspired by plan-and-solve prompting \citep{wang-etal-2023-plan} and the use of scratchpad for LLMs to show intermediate output in a specified region \citep{nye2021workscratchpadsintermediatecomputation}. We introduce the following XML tags to help guide LLMs on structuring their thoughts and responses as show in \ref{fig:thinking_format} (c). The \texttt{<thoughts></thoughts>} and \texttt{<response></response>} tags are the high-level tags and only exist once in each response. \texttt{<plan></plan>} only contains the \texttt{<step></step>} tags denoting each step, while the rest of the tags are utilized for each step.

The intuition behind designing the thinking process in this way is that, at each step, we encourage the model to first generate a title summarizing what should be done, followed by details or intermediate results recorded in the scratchpad. Then, we instruct the model to summarize its progress and determine the next step based on the plan. This design aligns with the plan-and-solve prompting approach, where prompting an LLM to generate a plan first can enhance its performance \citep{wang-etal-2023-plan}. A complete example of this structured thinking style is provided in \Cref{appendix:structured_example}.

\subsection{Data Preparation}

To prepare a training set, we select open datasets across five domains: 1) mathematics, 2) instruction following, 3) coding, 4) safety, and 5) finance. The rationale behind this mixture is that mathematics is known to improve STEM performance of LLMs \citep{olmo20252olmo2furious}, instruction following ensures generalization of reasoning capabilities, and code serves as a strong anchor language for an LM to think through and is known to improve performance \citep{gao2023palprogramaidedlanguagemodels,chen2023programthoughtspromptingdisentangling}. Finally, safety is included to improve the safety of the model, which may degrade from the introduction of long thinking, and finance is included to provide a representative of domain-specific data, which we will utilize to experiment and observe its effect on overall performance.

\subsubsection{Data Mixture}\label{sec:data_mixture}

The selected open datasets for each domain and their proportions are described in \Cref{tab:data_mix} in \Cref{appendix:data_mixture}. While the majority of the datasets provide a straightforward ground truth answer, PRM800K is different. 

PRM800K is a dataset used to construct a process reward model, which is a different type of model from an LLM. Therefore, we create a simple script to convert choices of each step in each ground truth into one complete response. At each step, we randomly select the correct or incorrect step to be included. In the case of an incorrect step, the script adds an additional step reflecting that the step is incorrect and resamples another step. This way, we create a dataset containing behaviors of self-correction, one of the behaviors common in reasoning models \citep{deepseekai2025deepseekr1incentivizingreasoningcapability}, in our training set.

We also perform post-processing on UltraFeedback by selecting only high-quality responses, filtering for records with a rating of the chosen response higher than 4.25. Finally, we downsample some large datasets through uniform sampling from their respective training splits. In total, the training set consists of \textbf{55,677 records}.

\subsubsection{Pipeline for Data Transformation and Refinement}\label{sec:data_pipeline}

Nevertheless, the training at this point only consists of a typical train-split fine-tuning set and does not include long reasoning chains like those demonstrated by reasoning models. Therefore, we prepare an LLM-based transformation-and-refinement pipeline for generating long reasoning chains from the training set. The pipeline consists of two main stages: 1) few-shot structured thinking transformation and 2) response refinement. In the first stage, we use few-shot prompting \citep{NEURIPS2020_1457c0d6} with GPT-4o mini (\texttt{gpt-4o-mini-2024-07-18}), along with ground truth associated with each query. There are a total of three few-shot exemplars (available in \Cref{appendix:few_shots}), which are manually curated by combining and structuring multiple state-of-the-art LLM responses.

The reason for transforming each ground truth into a structured thinking format first is to allow conversion into semi-structured and unstructured thinking formats later through simple text replacements (i.e., removing some or all auxiliary XML tags). This ensures equivalent levels of information, which is crucial for evaluating the effectiveness of different thinking formats.

After transformation, we utilize Qwen2.5-32B-Instruct\footnote{\url{https://huggingface.co/Qwen/Qwen2.5-32B-Instruct}} to refine the generated response, ensuring structural and informational correctness. We prompt the LLM to correct formatting issues, fill in missing content, and improve response quality based on generated thoughts. The same few-shot exemplars are used to guide this refinement process.

We derive semi-structured and unstructured thinking formats by systematically removing auxiliary XML tags. Specifically, we replace tags with words to maintain contextual consistency and strip away all structural elements except for \texttt{<thoughts></thoughts>} and \texttt{<response></response>} tags to create semi-structured data. In contrast, unstructured thinking data is obtained by removing or substituting all XML tags.

In total, we have approximately 67M tokens in our training set ready for SFT. The average number of tokens per instruction is 145.20, while the average number of tokens per output is 1,060.94, increased from the original 248.58 tokens. The maximum number of steps in the thinking part (excluding planning) is 24, with an average of 4.99 steps.

\subsection{Evaluation}
We utilize the following six benchmarks from different domains to assess the model's performance: 1) \textbf{GSM8K} \citep{cobbe2021trainingverifierssolvemath} is a math word problem dataset used to evaluate mathematical performance; 2) \textbf{HumanEval+} \citep{evalplus,evalperf}, which evaluates code generation, as the code domain is present in our training data; 3)\textbf{IFEval} \citep{zhou2023instructionfollowingevaluationlargelanguage} is selected to evaluate instruction-following performance; 4) \textbf{GPQA} \citep{rein2024gpqa} and 5) \textbf{MMLU Pro} \citep{wang2024mmlupro} are included as additional challenging QA benchmarks; and 6) \textbf{ThaiExam} \citep{pipatanakul2023typhoon} is a multiple-choice benchmark consisting of standard exams used to evaluate Thai students. We used this benchmark to evaluate the model's performance in Thai, given that the original model was tuned for improved Thai language capabilities.
 
\section{Experiments}\label{sec:experiments}
For all experiments in this section, we used the setup described in \Cref{appendix:experimental_setup}. We conduct five experiments, each presented in its own subsection, to answer the following five research questions.

\begin{enumerate}
    \item[Q1] \textit{Does the thinking format matter for reasoning models? If so, what is the best format?}
    \item[Q2] \textit{How much data is needed to train a reasoning model?}
    \item[Q3] \textit{In which domain of the dataset matters most when developing a reasoning model?}
    \item[Q4] \textit{How can a reasoning model be adapted to generate its reasoning trace in Thai?}
    \item[Q5] \textit{What are the differences when generating reasoning traces in Thai versus in English for a multilingual reasoning model?}
\end{enumerate}

\subsection{Structured Thinking Provides Improvements for Mathematics and Coding}\label{sec:thinking_formats}

To assess the impact of training a model on a long-thinking dataset and to evaluate the effects of various thinking formats, we designed six distinct scenarios. The results are presented in \Cref{tab:data_formats}:

\begin{itemize}
    \item \textbf{Typhoon 2 3B Instruct}: These scenarios evaluate the base model.
    \begin{enumerate}
        \item \textbf{Zero-shot}: We prompt the model in a zero-shot manner to establish its baseline performance. Note that modern LLMs may have been exposed to CoT data during training; consequently, the model might exhibit chain-of-thought behavior even in zero-shot settings.
        \item \textbf{Zero-shot CoT}: We prompt the LLM with the phrase ``Let's think step by step,'' as suggested by \citep{kojima2022large}, to observe its performance when explicitly encouraged to reason before answering, without any additional training.
        \item \textbf{SFT}: To compare standard supervised fine-tuning on a typical training set with training on a long-thinking dataset, we fine-tune the base model on a combined training split of non-long-thinking data and then evaluate its performance using zero-shot prompting.
    \end{enumerate}
    \item 4.-6. \textbf{Typhoon T}: This variant of Typhoon 2 3B Instruct is trained on a long-thinking dataset generated following the approach described in \Cref{sec:methods}. We experiment with three model variants—\textbf{Unstructured}, \textbf{Semi-structured}, and \textbf{Structured}—each trained on its respective thinking-formatted dataset.
\end{itemize}

\begin{table}[tbp]
\small
\centering
\begin{threeparttable}
\caption{Performance of models on each benchmark (higher is better). \textbf{Bold} indicates the best score in each column. \underline{Underlined} scores denote improvements over the baseline, Typhoon 2 3B Instruct. We apply this convention across all results tables. \textbf{Typhoon 2} refers to Typhoon 2 3B Instruct, and \textbf{Typhoon T} refers to its variant trained on a long-thinking dataset. \textbf{SFT} refers to supervised fine-tuning on the original datasets. All reasoning models show improvement over SFT on the original dataset.}
\label{tab:data_formats}
\begin{tabular}{lcccccc}
\toprule
\bf Model & \bf GSM8K & \bf HumanEval+ & \bf IFEval & \bf GPQA & \bf MMLU Pro & \bf ThaiExam \\
\midrule
\textbf{Typhoon 2} &  &  &  &  &  &  \\
\quad Zero-Shot & 57.32 & 63.51 & \textbf{69.32} & 25.00 & \textbf{26.61} & 32.69 \\
\quad Zero-Shot CoT         & 53.83 & 0.00  & 68.95        & \underline{25.45} & 23.36       & \underline{\textbf{33.27}} \\
\quad SFT           & 20.62 & 46.24 & 17.74        & 16.74  & 13.96 & 15.65 \\
\midrule
\textbf{Typhoon T} &  &  &  &  &  &  \\
\quad Unstructured            & \underline{59.82} & \underline{67.88} & 34.01        & 24.78 & 20.44 & 21.36 \\
\quad Semi-structured         & 57.24 & \underline{\textbf{72.87}} & 55.27 & \underline{\textbf{27.68}} & 19.46 & 21.92 \\
\quad Structured              & \underline{\textbf{62.02}} & \underline{69.76} & 53.60 & \underline{27.23} & 23.56 & 22.84 \\
\bottomrule
\end{tabular}
\end{threeparttable}
\vspace{-10pt}
\end{table}

We observe an unexpected result when applying zero-shot CoT prompting with the base model: a slight performance decrease across most benchmarks, except for GPQA and ThaiExam, where marginal improvements are observed. This finding contradicts the common observations that zero-shot CoT prompting enhances model performance by eliciting reasoning. We hypothesize that this decline may be due to additional training steps on Thai data, which have affected LLM's reasoning performance \citep{khade-etal-2025-challenges}. Furthermore, zero-shot CoT prompting led the model to generate completely incorrect code, scoring $0.00$, likely due to distractions from unhelpful reasoning traces.

Similarly, we observe a significant drop in performance when applying SFT with the non-long-thinking datasets. This suggests symptoms of catastrophic forgetting \citep{luo2025empiricalstudycatastrophicforgetting}, where the model shifts its distribution toward the fine-tuned data and loses generalization capabilities \citep{kotha2024understanding}. In contrast, SFT with long-thinking data does not exhibit the same behavior. Nevertheless, models fine-tuned on the long-thinking dataset show degraded performance in instruction-following tasks (IFEval). Additionally, ThaiExam scores decrease, which is expected given that the fine-tuning dataset primarily consists of English data, thereby reducing the model’s performance on Thai-language benchmarks.

We further analyzed the average response length on the MMLU Pro benchmark and found that Typhoon T models generally generate longer responses (more output tokens) than Typhoon 2 3B Instruct variants. This demonstrates the effectiveness of SFT on long-thinking datasets, except when prompting Typhoon 2 with zero-shot chain-of-thought reasoning, which generates long but unhelpful responses. Additional results and discussions are provided in \Cref{appendix:output_tokens}.

Overall, unstructured thinking appears to be suboptimal compared to other formats, while structured thinking yields improvements on a greater number of benchmarks. We attribute the superior performance of structured thinking to the presence of auxiliary separating tags, which help segment different reasoning components. Moreover, we observe that structured thinking results in more concise responses compared to semi-structured thinking, suggesting increased reasoning efficiency. This is likely due to the biases introduced by additional auxiliary structural tags.

Interestingly, we observe rare instances where the structured-thinking model, when prompted with a Thai query using non-zero temperature, generates its reasoning traces in Thai--despite not being trained on Thai long-thinking data. This suggests that the model may generalize its reasoning patterns across languages.

\subsection{Balancing Data Quantity: The Key to Optimal Reasoning Model Performance}\label{sec:data_sizes}
 
To evaluate the impact of data quantity on the performance of the reasoning model, we construct downsampled versions of the main training dataset. The baseline for this experiment is \textbf{Typhoon T}, trained on structured long-thinking data from the previous subsection. To create the downsampled datasets, we sample $x\%$ from each subdataset that comprises the full dataset, ensuring the overall data distribution is preserved. We consider a range of $x$ values: $\{75, 50, 25, 10, 5\}$.

\begin{wrapfigure}{r}{0.5\textwidth}
\vspace{-20pt}
  \begin{center}
    \includegraphics[width=0.48\textwidth]{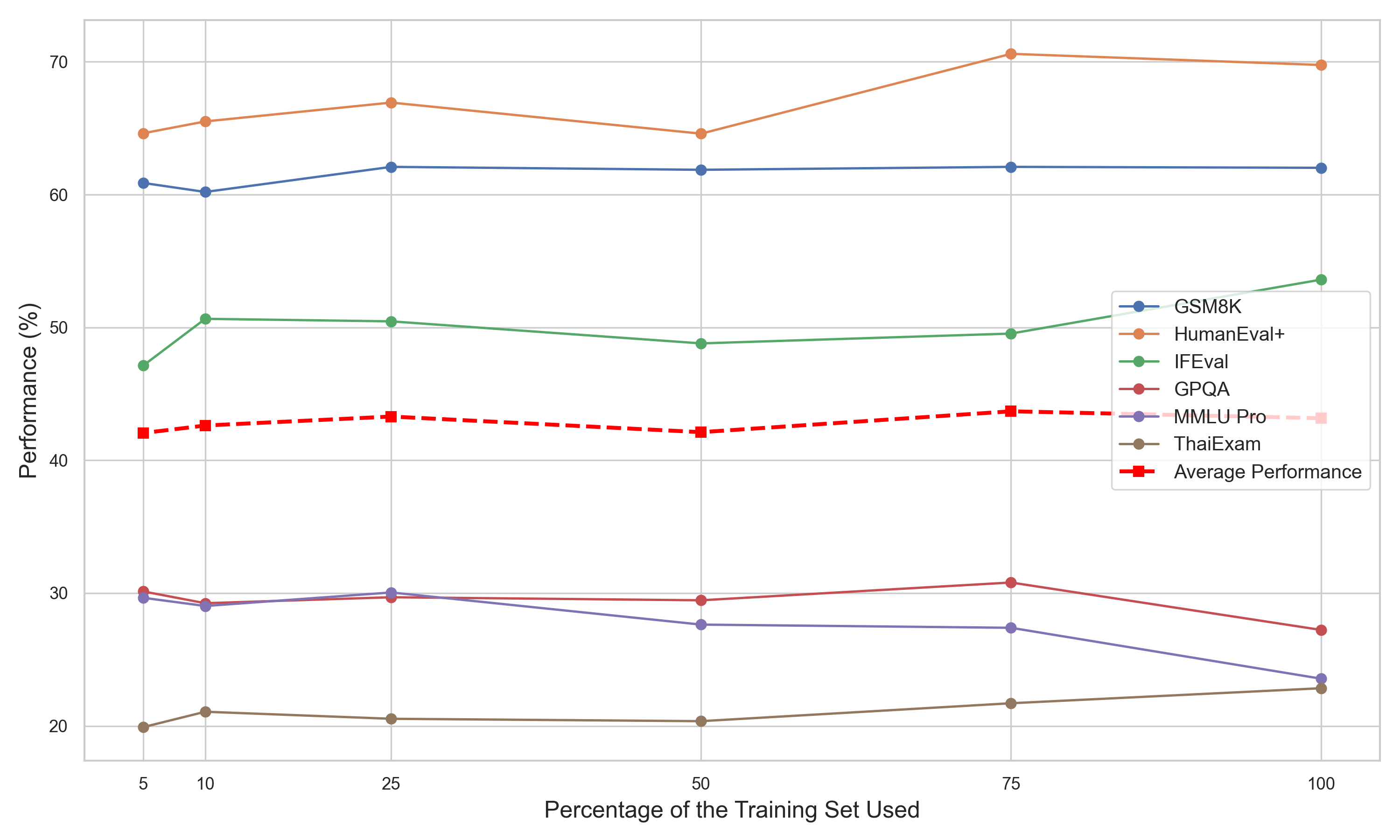}
  \end{center}
  \setlength{\belowcaptionskip}{-10pt}
  \caption{Increasing the proportion of the training set beyond 75\% results in performance degradation for some datasets, while GSM8K generally shows a trend of performance improvement with the proportion.}
  \label{fig:training_percentage}
\end{wrapfigure}

We note that we fixed the number of epochs rather than the total training steps, meaning that models trained on smaller datasets undergo fewer optimization steps. This ensures that each subset is trained for a proportional duration relative to the full dataset without artificially extending training. If we had controlled for total training steps instead (e.g., by increasing epochs for smaller datasets), we might observe different trends, potentially increasing the risk of overfitting. The results of this experiment are presented in \Cref{fig:training_percentage}. The full results in a table format is available in \Cref{appendix:data_sizes}.

The findings suggest that an excessive amount of data may not be optimal for achieving the best performance. However, reducing the dataset size too much also leads to a decline in instruction-following performance (IFEval). Similarly, MMLU Pro exhibits a downward trend in performance beyond a certain data threshold, 50\% in this experiment.

In contrast, HumanEval+ benefits from an increase in training data. Among the tested configurations, training on 75\% of the total dataset (41,755 records) yielded the most effective structured thinking reasoning model. We designate the model trained on this dataset as \textbf{Typhoon T1-EN}.

\subsection{Don't Leave Safety Out When Training a Reasoning Model}\label{sec:loo}

To assess the impact of each domain in the training set, we perform a leave-one-out experiment, where we train a model following the approach in \Cref{sec:thinking_formats} on a training dataset with one specific domain removed. We use the best-performing configuration from the previous subsection (75\% dataset), which also serves as our baseline. We expect to observe a decrease in performance and measure the impact of each domain by the extent of this decrease. The full results and additional discussions are available in \Cref{appendix:loo}.

Among all domains, the safety domain has the most impact on performance. Removing it results in substantial performance drops in GSM8K, HumanEval+, IFEval, and ThaiExam. This effect can be partly attributed to the role of the safety domain in aligning the model to be helpful \citep{wang2023helpsteermultiattributehelpfulnessdataset}. Without this alignment, the model's overall capabilities decline across multiple benchmarks. Interestingly, the model performs the same or better on challenging English multiple-choice benchmarks, such as GPQA and MMLU Pro.

In summary, our findings emphasize the necessity of a diverse training set—including prompt diversity, response style diversity, and domain diversity. While instruction-following is useful, it may not be as crucial as previously assumed for training a model to generalize its reasoning behavior across prompts. Instead, safety-focused instruction-following datasets prove to be more beneficial.

\subsection{Training with Thai-Translated Data Improves Thai Performance at the Cost of Others}\label{sec:thai_thinking}

To equip the model with the ability to generate Thai reasoning traces--especially when the user's prompt is in Thai or explicitly requests Thai output--we train a model with additional translated Thai content. To obtain this translated content, we construct a translation pipeline described in \Cref{appendix:translation_pipeline}.

In this subsection, we investigate the best approach for equipping the reasoning model with the ability to generate Thai reasoning traces. We experiment with two main approaches: (1) continual SFT, where we take Typhoon T1-EN and perform an additional SFT with the translated Thai dataset, and (2) SFT from scratch, where we train from the beginning using a mixture of the translated Thai dataset and the Typhoon T1-EN training data. We found that the continual SFT approach yielded subpar performance compared to SFT from scratch, as shown in \Cref{tab:thai_perf}.

\begin{table}[tbp]
\centering
\small
\begin{threeparttable}
\caption{Performance of model variants on various benchmarks, evaluating the impact of additional training data. \textbf{CSFT} refers to continual SFT. Adding 1.5k samples improves IFEval and ThaiExam scores, while CSFT significantly reduces overall performance.}
\label{tab:thai_perf}
\begin{tabular}{lcccccc}
\toprule
\bf Model & \bf GSM8K & \bf HumanEval+ & \bf IFEval & \bf GPQA & \bf MMLU Pro & \bf ThaiExam \\
\midrule
\textbf{Typhoon T1-EN} & \textbf{62.09} & \textbf{70.60} & 49.54 & \textbf{30.80} & \textbf{27.39} & 21.71 \\
\quad + 1.5k, CSFT & 41.39 & 65.79 & 33.83 & 23.66 & 4.30 & 21.20 \\
\quad + 1.5k & 60.12 & 67.90 & \underline{\textbf{51.76}} & 29.91 & 19.32 & \underline{\textbf{23.56}} \\
\quad + 1k & 61.94 & 66.77 & \underline{50.09} & 24.55 & 23.48 & 21.57 \\
\quad + 0.5k & 60.88 & 68.24 & \underline{49.72} & 25.45 & 23.05 & \underline{22.62} \\
\bottomrule
\end{tabular}
\end{threeparttable}
\vspace{-10pt}
\end{table}

To better understand the effect of incorporating Thai-translated content, we experimented with different proportions of the translated dataset: using 2/3 and 1/3 of the full translated dataset, approximately 1,000 and 500 records, respectively. The results are presented in \Cref{tab:thai_perf}.

We observe that all models trained with additional Thai-translated data exhibit improved instruction-following capabilities (IFEval), with the most improvement occurring when training on the full Thai-translated dataset. Similarly, performance on ThaiExam follows this trend, showing improvements as more Thai training data is included. However, we also observe a decline in performance on other benchmarks, likely indicating a trade-off due to the inherent capacity limitations of the model’s 3B parameters \citep{li-etal-2024-teaching,allen-zhu2025physics}.

We observe that the models trained with Thai translated data are able to generate its thinking trace in Thai when prompted in Thai (see \Cref{appendix:thai_thinking}), while maintaining performance similar to that of Typhoon T1-EN. The model that achieved the highest improvement on ThaiExam--trained with the full Thai-translated dataset--is named \textbf{Typhoon T1}.

In addition, we observe that on ThaiExam, the percentage of Thai characters in the thinking trace (i.e., the content between \texttt{<thoughts>} tags) increased from 16.67\% when generated with Typhoon T1-EN to 95.49\%. This increase in the number of Thai characters, along with improved performance on ThaiExam, demonstrates the effectiveness of training with only a small portion of Thai-translated data. We hope this approach sparks interest in further studies on equipping reasoning models with the ability to generate their reasoning traces in low-resource languages.

\subsection{Let The Reasoning Model Chooses Its Own Reasoning Language}\label{sec:force_thinking}

To better understand the effect of thinking language on the performance, we conduct an experiment using \textbf{Typhoon T1} from the previous section. Specifically, we examine the performance implications of enforcing thought generation in either English or Thai through prompting. This differs from a zero-shot approach, used in the previous subsection, where we allow the model to choose its own reasoning language--typically English for English prompts and Thai for Thai prompts.

We employ a standardized user prompt, provided in \Cref{appendix:force_prompt}, to explicitly direct \textbf{Typhoon T1} to generate its reasoning traces in either English or Thai before producing a final answer. The results, presented in \Cref{tab:force_lang}, indicate that restricting the model’s reasoning language negatively impacts overall task performance across multiple benchmarks. Notably, while both forced English and Thai reasoning lead to performance degradation, English slightly outperforms Thai in most cases.

\begin{table}[tbp]
\small
\centering
\begin{threeparttable}
\caption{\textbf{EN} denotes forced reasoning in English, and \textbf{TH} denotes forced reasoning in Thai. Constraining Typhoon T1 to reason in a specific language degrades overall accuracy. English reasoning is slightly more effective than Thai reasoning across most benchmarks. However, allowing the model to choose its own thinking language yields the best performance.}
\label{tab:force_lang}
\begin{tabular}{lcccccc}
\toprule
\textbf{Model} & \textbf{GSM8K} & \textbf{HumanEval+} & \textbf{IFEval} & \textbf{GPQA} & \textbf{MMLU Pro} & \textbf{ThaiExam} \\
\midrule
\textbf{Typhoon T1}   & 60.12 & 67.90 & \textbf{51.76} & \textbf{29.91} & \textbf{19.32} & 23.56 \\
+ EN        & 46.17 & 0.00     & 48.98     & 26.56     & 16.55     & \underline{\textbf{25.31}}  \\
+ TH        & 48.29 & 0.00     & 44.73     & 25.67     & 16.05     & 24.66  \\
\bottomrule
\end{tabular}
\end{threeparttable}
\vspace{-10pt}
\end{table}

Our findings suggest that constraining the reasoning language of a multilingual model may disrupt its ability to leverage cross-linguistic representations effectively. The observed performance degradation when enforcing English or Thai reasoning highlights the model's reliance on flexible multilingual processing for optimal results. This aligns with prior research indicating that multilingual models often synthesize knowledge across languages rather than strictly adhering to a single linguistic paradigm \citep{chen-etal-2024-breaking,hua-etal-2024-mothello}. One possible explanation for the superior performance of unconstrained reasoning is that Typhoon T1 dynamically selects the most effective linguistic structures for intermediate thought generation, drawing from both English and Thai as needed, similar to \cite{huang-etal-2023-languages,huang2025adacotrethinkingcrosslingualfactual}. Furthermore, the zero accuracy observed in HumanEval+ for both forced-language settings suggests that forcing a language can lead to the generation of unhelpful reasoning traces, similar to what was observed in \Cref{sec:thinking_formats}.

\subsection{Summary: Typhoon T1-EN And Typhoon T1}\label{sec:summary}

\begin{wrapfigure}{r}{0.5\textwidth}
\vspace{-20pt}
  \begin{center}
    \includegraphics[width=0.48\textwidth]{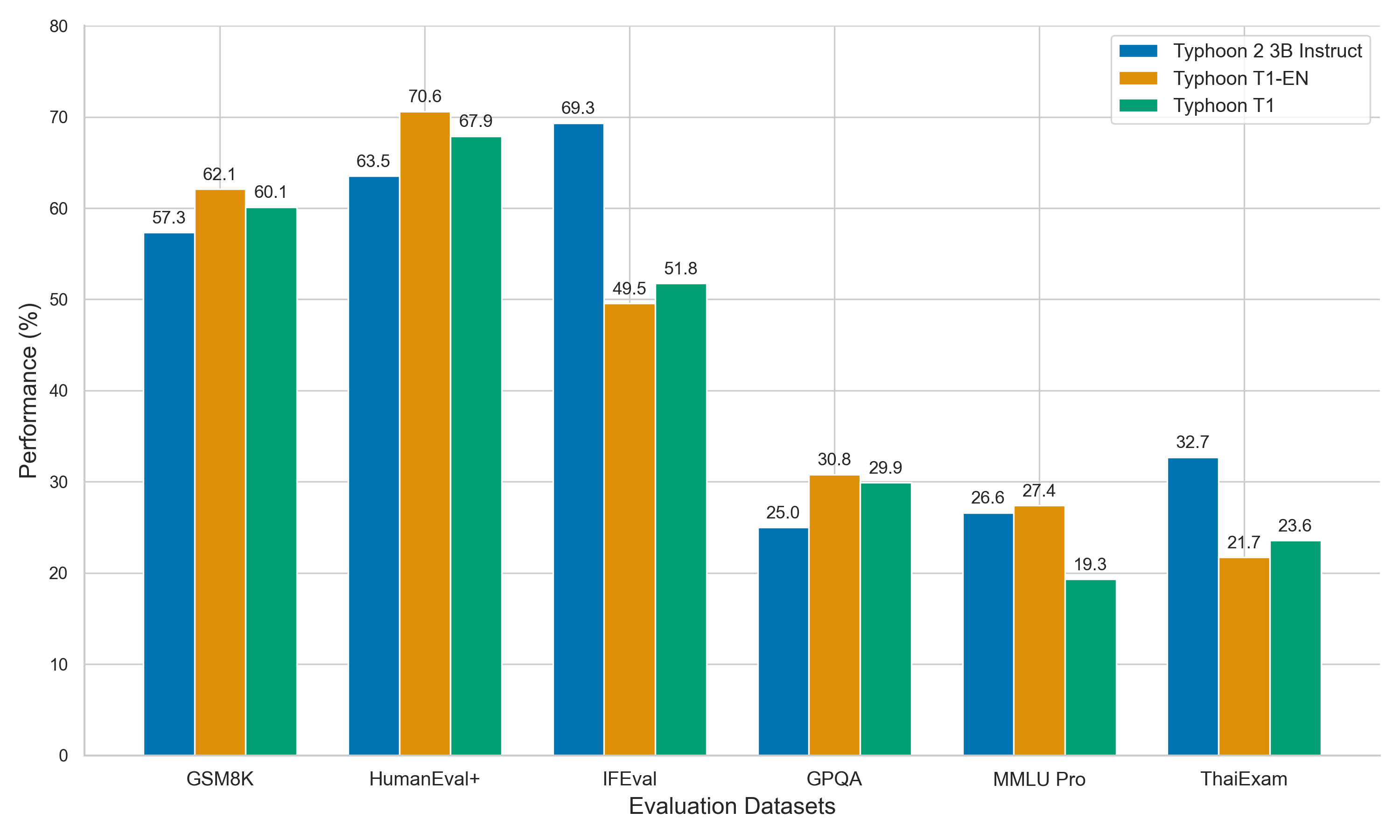}
  \end{center}
  \setlength{\belowcaptionskip}{-10pt}
  \caption{Final performance comparison of Typhoon T1-EN and Typhoon T1 against the baseline Typhoon T1 3B Instruct model across six evaluation benchmarks.}
  \label{fig:final_comparison}
\end{wrapfigure}

We compare the final performance of each model against Typhoon T1 3B Instruct in \Cref{fig:final_comparison}. Based on all experiments in this section, we found the most optimal configuration for training 3B-parameter reasoning models using the SFT approach as follows:

\begin{itemize}
    \item \textbf{Typhoon T1-EN}: Uses structured thinking with 41,755 records, trained using the standard settings described in \Cref{appendix:experimental_setup}.
    \item \textbf{Typhoon T1}: Extends Typhoon T1-EN's approach by incorporating an additional training set of 1,565 Thai-translated structured long-thinking records, as described in \Cref{sec:thai_thinking}. This enables the model to generate Thai reasoning traces while maintaining Thai performance.
\end{itemize}

\section{Conclusion}

We present an open recipe for developing the reasoning model \textbf{Typhoon T1}, which requires no distillation from other models, generalizes across domains, and produces reasoning traces in English or Thai. We also introduce structured thinking--using auxiliary XML tags to create efficient reasoning formats with fewer tokens. Although Typhoon T1 shows strong improvements on GSM8K, HumanEval+, and GPQA, our approach involves trade-offs in performance due to model size, such as reduced instruction following and Thai performance. We hope our insights on data size, domain, and reasoning language and artifacts accelerate research on reasoning models.


%

\bibliography{references}
\bibliographystyle{iclr2025_conference}

\appendix
\section{Appendix}

\subsection{Limitations and Future Work}

Due to computational constraints, we experiment with a relatively small model, consisting of 3B parameters. This introduces inherent limitations in the model’s reasoning capabilities and trade-offs in its performance \citep{li-etal-2024-teaching,allen-zhu2025physics}. However, our approach demonstrates promising improvements in reasoning ability through structured long-thinking approach. Additionally, we restrict our experiments to zero-shot prompting and do not incorporate test-time scaling techniques, such as those introduced in \cite{snell2024scalingllmtesttimecompute}, which could further enhance performance.

Several important properties remain unexplored in our study of small reasoning models, including the impact of the number of reasoning steps, variations in structured thinking pattern, and a deeper analysis on finer-grained reasoning behaviors such as self-correction and problem decomposition. In addition, we did not conduct a comparison with a reinforcement learning baseline, given obscure details and inconclusive standards on this approach at the time of writing. Future work should investigate these aspects to better understand their contributions to model reasoning capabilities. 

\subsection{Related Work}
\subsubsection{Chain-of-Thought Prompting}

Chain-of-Thought (CoT) prompting \citep{wei2022chain} is an approach used to elicit reasoning in LLMs through in-context reasoning exemplars \citep{NEURIPS2020_1457c0d6}. This approach improves the performance of LLMs by generating additional tokens--``thoughts''--preceding a final answer. Similarly, zero-shot CoT prompting \citep{kojima2022large} elicits the generation of a reasoning chain before the final answer without any in-context exemplars, relying instead on a simple phrase like ``Let\'s think step by step.''

Various extensions of CoT prompting \citep{yao2023tree, Besta_Blach_Kubicek_Gerstenberger_Podstawski_Gianinazzi_Gajda_Lehmann_Niewiadomski_Nyczyk_Hoefler_2024, zhou2023threadthoughtunravelingchaotic} have been proposed to further extend the reasoning process, allowing LLMs additional time to explore alternatives, verify intermediate responses, and correct themselves before providing a final answer. However, the complexity of these approaches increases significantly as they involve additional components such as verifiers and memory mechanisms for maintaining state.

\subsubsection{Test-Time Compute Scaling}

Test-time compute scaling \citep{snell2024scalingllmtesttimecompute} refers to allocating a greater compute budget during inference, allowing LLMs to engage in more extensive reasoning before generating a final answer. Several methods exist to achieve this, most of which involve search-based techniques, such as Monte Carlo Tree Search (MCTS) \citep{ding-etal-2024-everything,duan2025promptbasedmontecarlotree}, tree traversal \citep{yao2023tree,Besta_Blach_Kubicek_Gerstenberger_Podstawski_Gianinazzi_Gajda_Lehmann_Niewiadomski_Nyczyk_Hoefler_2024,bi2025forestofthoughtscalingtesttimecompute}, and other search strategies \citep{snell2024scalingllmtesttimecompute,wang2024openropensourceframework}. These methods are often paired with a reward model, either an outcome reward model \citep{beeching2024scalingtesttimecompute} or a process reward model \citep{snell2024scalingllmtesttimecompute}. Some studies \citep{qin2024o1replicationjourneystrategic,guan2025rstarmathsmallllmsmaster} suggest that reasoning models can be enhanced by fine-tuning LLMs on search traces. However, we find this approach computationally expensive, especially at scale. Our approach, in contrast, is simpler to implement and more cost-effective than the aforementioned methods.

\subsubsection{Reasoning Models}

Reasoning models, also referred to as ``thinking LLMs'' \citep{wu2024thinkingllmsgeneralinstruction}, represent a recent advancement in the field of large language models. These models have demonstrated effectiveness in solving complex benchmarks by generating extended reasoning chains, often scaling with problem difficulty \citep{deepseekai2025deepseekr1incentivizingreasoningcapability}. Reasoning models are typically developed by augmenting an LLM through additional supervised fine-tuning (SFT) \citep{muennighoff2025s1simpletesttimescaling} or reinforcement learning (RL) \citep{deepseekai2025deepseekr1incentivizingreasoningcapability}.

Nevertheless, many works \citep{skyt12025,bespokestratos,huang2024o1replicationjourney} have also demonstrated the effectiveness of knowledge distillation from a reasoning model. However, this distillation-based approach provides only a shortcut and requires access to an existing reasoning model. To gain a deeper understanding of reasoning models, we argue that developing such models from scratch--without leveraging an existing reasoning model--is a more effective approach. This study adopts that perspective to gain better insights into reasoning models.

\subsection{Data Mixture}\label{appendix:data_mixture}
\begin{figure}
    \centering
    \includegraphics[width=0.7\linewidth]{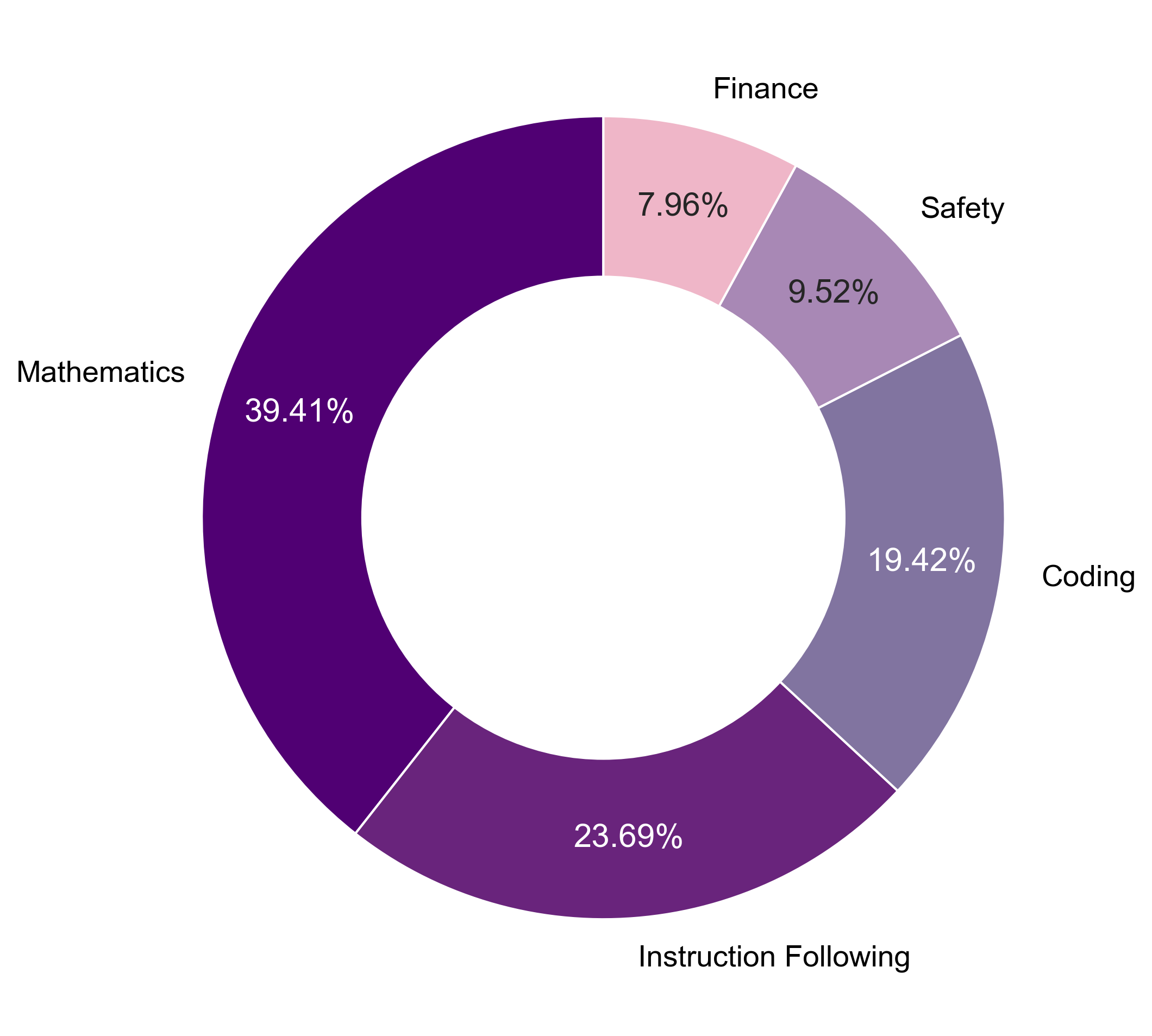}
    \caption{This figures show domain distribution of the training set for the experiments.}
    \label{fig:domain_dis}
\end{figure}

\Cref{tab:data_mix} shows data mixture of the training set used in the experiments in \Cref{sec:experiments}. \Cref{fig:domain_dis} shows data distribution of each domain in percentage.

\begin{table}[tbp]
\caption{Data mixture of the training set.}
\label{tab:data_mix}
\begin{center}
\begin{tabular}{lr}
\toprule
\bf Domain/Dataset & \bf \#Records \\
\midrule
\textit{Mathematics} & \textit{21,941} \\
\quad MATH \citep{hendrycks2021measuring}             & 7,500  \\
\quad Tulu 3 SFT Personas Math Grade \citep{lambert2025tulu3pushingfrontiers}     & 7,497  \\
\quad PRM800K Phase 2 \citep{lightman2024lets}   & 5,809  \\
\quad PRM800K Phase 1 \citep{lightman2024lets}   & 808    \\
\quad O1 Journey \citep{qin2024o1replicationjourneystrategic}        & 327    \\
\midrule
\textit{Instruction Following} & \textit{13,188} \\
\quad No Robots \citep{no_robots}        & 9,500  \\
\quad UltraFeedback \citep{cui2024ultrafeedbackboostinglanguagemodels}     & 3,688  \\
\midrule
\textit{Coding} & \textit{10,814} \\
\quad Evol codealpaca v1 \citep{luo2023wizardcoder}    & 5,564  \\
\quad Tulu 3 SFT Personas Code \citep{lambert2025tulu3pushingfrontiers}       & 5,250  \\
\midrule
\textit{Safety} & \textit{5,300} \\
\quad HelpSteer \citep{wang2023helpsteermultiattributehelpfulnessdataset}         & 5,300  \\
\midrule
\textit{Finance} & \textit{4,434} \\
\quad Wealth Alpaca \citep{wealth_alapca}            & 4,434  \\
\midrule
\bf Total & \bf 55,677 \\
\bottomrule
\end{tabular}
\end{center}
\end{table}

\subsection{An Example of Structured Thinking}\label{appendix:structured_example}

Given the following prompt: ``A model of a park was built on a scale of 1.5 centimeters to 50 meters. If the distance between two trees in the park is 150 meters, what is this distance on the model?'', we can expect to see the following response (indentation is manually included for readability):

\begin{lstlisting}[frame=single]
<thoughts>

<plan>
  <step order="1">Understanding the scale relationship</step>
  <step order="2">Setting up the proportion</step>
  <step order="3">Cross-multiplying to solve for the model distance</step>
  <step order="4">Calculating the final result</step>
  <step order="5">Verifying the calculations</step>
</plan>

<step>
  <title>Understanding the scale relationship</title>
  <scratch_pad>
  The scale given is 1.5 cm to 50 m. This means that every 1.5 cm on the model corresponds to 50 meters in reality.
  </scratch_pad>
  <summary>I'm clarifying the scale relationship to ensure accurate calculations.</summary>
  <next_step>Setting up the proportion</next_step>
</step>

<step>
  <title>Setting up the proportion</title>
  <scratch_pad>
  Let \( x \) be the distance on the model (in centimeters). The proportion can be set up as follows:
  \[
  \frac{1.5 \text{ cm}}{50 \text{ m}} = \frac{x \text{ cm}}{150 \text{ m}}
  \]
  This represents the relationship between the model and the actual distance.
  </scratch_pad>
  <summary>I've established the proportion based on the scale provided.</summary>
  <next_step>Cross-multiplying to solve for the model distance</next_step>
</step>

<step>
  <title>Cross-multiplying to solve for the model distance</title>
  <scratch_pad>
  Cross-multiplying gives us:
  \[
  1.5 \cdot 150 = 50 \cdot x
  \]
  This simplifies to:
  \[
  225 = 50x
  \]
  </scratch_pad>
  <summary>I've cross-multiplied to form an equation that allows solving for \( x \).</summary>
  <next_step>Calculating the final result</next_step>
</step>

<step>
  <title>Calculating the final result</title>
  <scratch_pad>
  Now, I will solve for \( x \):
  \[
  x = \frac{225}{50}
  \]
  Calculating this gives:
  \[
  x = 4.5 \text{ cm}
  \]
  </scratch_pad>
  <summary>I've calculated the model distance based on the proportion derived earlier.</summary>
  <next_step>Verifying the calculations</next_step>
</step>

<step>
  <title>Verifying the calculations</title>
  <scratch_pad>
  To verify:
  - The original scale is 1.5 cm to 50 m, which means 1.5 cm on the model should equal 50 m in real life.
  - The calculation shows that 4.5 cm on the model corresponds to 150 m in reality, which checks out with the scale.
  </scratch_pad>
  <summary>I've double-checked the calculations to ensure accuracy.</summary>
  <next_step></next_step>
</step>

</thoughts>

<response>
To find the distance between the two trees on the model, we use the scale provided.

### **Given:**
- Scale: 1.5 cm corresponds to 50 m.
- Distance between the trees in reality: 150 m.

### **Step-by-step Calculation:**

1. **Set up the proportion:**
   \[
   \frac{1.5 \text{ cm}}{50 \text{ m}} = \frac{x \text{ cm}}{150 \text{ m}}
   \]

2. **Cross-multiply:**
   \[
   1.5 \cdot 150 = 50 \cdot x
   \]
   Simplifying gives:
   \[
   225 = 50x
   \]

3. **Solve for \( x \):**
   \[
   x = \frac{225}{50} = 4.5 \text{ cm}
   \]

### **Final Answer:**
The distance between the two trees on the model is **4.5 centimeters**.
</response>
\end{lstlisting}

\subsection{Few-shot Exemplars for Data Transformation and Refinement}\label{appendix:few_shots}

\begin{itemize}
    \item \url{https://pastes.io/example-1}
    \item \url{https://pastes.io/example-2}
    \item \url{https://pastes.io/example-3}
\end{itemize}

\subsection{Experimental Setup}\label{appendix:experimental_setup}
We employ the following setup and tools for all our experiments. We use vLLM \citep{kwon2023efficient} for efficient inference in our data pipeline and \texttt{LlamaFactory} \citep{zheng2024llamafactory} for efficient multi-GPU training. Models are trained mostly on 2xH100s.

We perform a standard SFT on the training set using the same training configurations for all experiments. We save and evaluate models every 200 steps, with 5\% of the training set used as a validation set, and load the best model at the end. We train each model for two epochs with a learning rate of 2e-5 using a cosine scheduler without warmup steps. The batch size is set to 6 with 6 gradient accumulation steps. For efficient training, we leverage \texttt{DeepSpeed ZeRO} Stage 2 \citep{9355301}, \texttt{Flash Attention 2} \citep{dao2023flashattention2fasterattentionbetter}, and \texttt{Liger Kernel} \citep{hsu2025ligerkernelefficienttriton} with \texttt{bfloat16} precision. The training configuration is available in \Cref{appendix:training_config}.

For evaluation, we use the open-source evaluation platform \texttt{olmes} \citep{gu2024olmesstandardlanguagemodel} for GSM8K, HumanEval+, IFEval, GPQA, and MMLU Pro. We use the default evaluation configuration for all experiments, except for modifying the maximum context length to approximately 16,384 tokens to allow reasoning models to think longer. For ThaiExam, we use our simple evaluation program that applies regular expressions to extract answers and evaluates using exact match criteria. The \texttt{olmes}'s evaluation configuration is available in \Cref{appendix:eval_config}.

\subsubsection{Training Configuration For LlamaFactory}\label{appendix:training_config}

\begin{lstlisting}[frame=single]
### model
model_name_or_path: scb10x/llama3.2-typhoon2-3b-instruct

### method
stage: sft
do_train: true
finetuning_type: full
deepspeed: examples/deepspeed/ds_z2_config.json

### dataset
dataset: <dataset_name>
template: llama3
overwrite_cache: true
preprocessing_num_workers: 16

### output
output_dir: outputs/<outupt_model_path>
logging_steps: 10
save_steps: 200
plot_loss: true
overwrite_output_dir: true

### train
per_device_train_batch_size: 6
gradient_accumulation_steps: 6
learning_rate: 2.0e-5
num_train_epochs: 2.0
lr_scheduler_type: cosine
warmup_ratio: 0.0
bf16: true
ddp_timeout: 180000000
use_unsloth: false
enable_liger_kernel: true
load_best_model_at_end: true
resume_from_checkpoint: false
flash_attn: fa2

### eval
val_size: 0.05
per_device_eval_batch_size: 1
eval_strategy: steps
eval_steps: 200
\end{lstlisting}

\subsubsection{Evaluation Configuration For \texttt{olmes}}\label{appendix:eval_config}

\begin{lstlisting}[frame=single]
olmes --model <model_name> --task gsm8k::olmes codex_humanevalplus::tulu ifeval::tulu gpqa::llama3 mmlu_pro:cot::none --model-type vllm --output-dir results/<model_name>
\end{lstlisting}

\subsection{Average Output Tokens}\label{appendix:output_tokens}

This section presents the average number of output tokens generated by each model on the listed benchmarks. The results, obtained from the experiment described in \Cref{sec:thinking_formats}.

\begin{table}[tbp]
\small
\centering
\begin{threeparttable}
\caption{Average number of output tokens generated by each model on the benchmarks.}
\label{tab:output_tokens}
\begin{tabular}{lcccc}
\toprule
\bf Model & \bf GSM8K & \bf GPQA & \bf MMLU Pro & \bf ThaiExam \\
\midrule
\textbf{Typhoon 2} &  &  &  &  \\
\quad Zero-shot             & 104.61 & 384.78  & 130.41 & 21.90  \\
\quad Zero-shot CoT        & 741.97 & 1238.54 & 1697.96 & 149.19 \\
\quad SFT & 72.22  & 479.55  & 91.25  & 587.95 \\
\midrule
\textbf{Typhoon T} &  &  &  &  \\
\quad Unstructured          & 169.03 & 478.53  & 491.33 & 829.21 \\
\quad Semi-structured       & 170.20 & 795.38  & 487.39 & 900.90 \\
\quad Structured            & 102.96 & 466.21  & 293.23 & 995.04 \\
\bottomrule
\end{tabular}
\end{threeparttable}
\end{table}

The results in \Cref{tab:output_tokens} reveal notable variations in the average output tokens across different model configurations and benchmarks. In particular, when prompted with zero-shot CoT phrases, the Typhoon 2 3B Instruct produces a substantially higher number of tokens—especially on GPQA and MMLU Pro—suggesting that the model is distracted by unhelpful chain-of-thought reasoning and achieves poor results.

In contrast, the supervised fine-tuned Typhoon 2 3B Instruct on non-long-thinking datasets generates more concise outputs on GSM8K and MMLU Pro, reflecting the style of the training set, while exhibiting a significant increase in output length on ThaiExam. The observation on ThaiExam is likely due to the training set not containing Thai records, making it out-of-training-distribution.

\subsection{Data Size Ablation Study Results}\label{appendix:data_sizes}

\Cref{tab:data_sizes} shows a full result of the experiment in \Cref{sec:data_sizes}.

\begin{table}[tbp]
\caption{Performance at different dataset sizes. Smaller dataset sizes can sometimes outperform the 100\% baseline, particularly in GPQA and MMLU Pro.}
\label{tab:data_sizes}
\centering
\begin{threeparttable}
\begin{tabular}{lcccccc}
\toprule
\bf Dataset Size & \bf GSM8K & \bf HumanEval+ & \bf IFEval & \bf GPQA & \bf MMLU Pro & \bf ThaiExam \\
\midrule
100\% & 62.02 & 69.76 & \textbf{53.60} & 27.23 & 23.56 & \textbf{22.84} \\
75\%  & \underline{\textbf{62.09}} & \underline{\textbf{70.60}} & 49.54 & \underline{\textbf{30.80}} & \underline{27.39} & 21.71 \\
50\%  & 61.87 & 64.59 & 48.80 & \underline{29.46} & \underline{27.63} & 20.36 \\
25\%  & \underline{\textbf{62.09}} & 66.93 & 50.46 & \underline{29.69} & \underline{\textbf{30.05}} & 20.54 \\
10\%  & 60.20 & 65.51 & 50.65 & \underline{29.24} & \underline{29.03} & 21.07 \\
5\%   & 60.88 & 64.62 & 47.13 & \underline{30.13} & \underline{29.65} & 19.91 \\
\bottomrule
\end{tabular}
\end{threeparttable}
\end{table}

\subsection{Leave-of-Out Experiment Results}\label{appendix:loo}

\Cref{tab:loo_experiment} shows a full result of the experiment in \Cref{sec:loo}.

\begin{table}[tbp]
\caption{Leave-one-out experiment results, assessing the impact of removing specific domains from training. \textcolor{red}{\textit{red}} values highlight the largest performance drop in each column. The ``-'' symbol denotes the removal of the corresponding domain from training. Excluding mathematical reasoning strongly improves IFEval performance, while safety removal boosts MMLU Pro.}
\label{tab:loo_experiment}
\centering
\begin{threeparttable}
\small
\begin{tabular}{lcccccc}
\toprule
\bf Model & \bf GSM8K & \bf HumanEval+ & \bf IFEval & \bf GPQA & \bf MMLU Pro & \bf ThaiExam \\
\midrule
\textbf{Typhoon T1-EN}           & \textbf{62.09} & \textbf{70.60} & 49.54 & \textbf{30.80} & 27.39 & 21.71 \\
\quad - IF          & 59.59        & 69.57        & 46.58        & 29.02        & 26.34     & \textbf{22.64}     \\
\quad - Math        & 59.51        & 69.47        & \underline{\textbf{53.60}} & \textcolor{red}{\textit{25.45}}        & \underline{28.52}     & 20.88     \\
\quad - Code        & 56.94        & \textcolor{red}{\textit{64.24}}        & 41.96        & 27.68        & \underline{27.65}     & 19.57     \\
\quad - Safety      & \textcolor{red}{\textit{56.71}}        & 64.35        & \textcolor{red}{\textit{41.59}}        & 30.13        & \textbf{29.38}     & \textcolor{red}{\textit{17.19}}     \\
\quad - Finance     & 61.94        & 67.06        & \underline{50.65}        & 27.90        & \textcolor{red}{\textit{20.45}}     & 18.68     \\
\bottomrule
\end{tabular}
\end{threeparttable}
\end{table}

Removing instruction-following results in a slight performance decrease across all benchmarks, except for ThaiExam. This outcome is expected, as instruction-following is a foundational skill for an LLM to perform well across tasks, since most tasks require a model to understand instructions to effectively utilize its domain knowledge. We note that the slight increase in ThaiExam performance may be due to the model being exposed to more English instruction-following data, thereby mitigating the impact of Typhoon fine-tuning on Thai data \citep{pipatanakul2024typhoon2familyopen}.

Excluding the mathematics domain from the training set leads to a decline in GSM8K scores, a mathematics-focused benchmark. However, we observe a notable increase in the mathematics subject of MMLU Pro, rising from 23.69\% to 29.90\%. A key difference between MMLU Pro and the other benchmarks is that MMLU Pro consists of multiple-choice questions.

One possible explanation for this phenomenon is the \textit{distribution shift}. Since our training set primarily contains mathematics problems requiring numerical responses rather than multiple-choice answers, removing the mathematics domain may allow the model to better adapt to multiple-choice formats. This finding highlights the importance of data diversity--not only in topics but also in response formats.

Unexpectedly, removing the coding domain negatively affects performance across all datasets. This finding aligns with previous research suggesting that coding serves as a strong representation \citep{gao2023palprogramaidedlanguagemodels} and enhances reasoning capabilities in LLMs \citep{chen2023programthoughtspromptingdisentangling}. We also observe a significant performance drop in IFEval, suggesting that logical instruction-following capabilities may be partially induced through exposure to coding datasets.

Finally, removing domain-specific datasets--finance-related data in this case--generally leads to slight performance drops across all benchmarks and a notable decrease in MMLU Pro. The decline is particularly evident in the economics subject of MMLU Pro, where accuracy drops from 40.05\% to 25.24\%. This result underscores the importance of domain-specific datasets when training a reasoning model using supervised fine-tuning.

\subsection{Translation Pipeline}\label{appendix:translation_pipeline}

The translation pipeline consists of an LLM, a fine-tuned Llama 3.1 8B Instruct in our case, prompted to act as a translator while adhering to specific rules, such as preserving structural XML tags. The exact prompt used for this purpose is provided bekiw. We translated approximately 25,000 records from the original 100\% training set of the Typhoon T structured long-thinking model referenced in \Cref{sec:thinking_formats}. We applied strict post-processing criteria, including removing prefixes that were not part of the translated content and filtering out translated instructions containing XML tags. After this process, we retained 1,565 high-quality translated records.

\begin{lstlisting}[frame=single]
Translate the following content into Thai appropriately.

**Rules**:
- Do not translate the tags used for structural formatting. Keep all XML tags intact and include in the final response.
- Translate only the content in English. Do not translate any other languages.
- Consider the context and adjust accordingly. You may change the order or add additional content to ensure fluency and make the text sound natural.
- Generate only the translated content, without explanations or formatting tags.
- Translate all content, not just the response tags.
- Do not change to a new structure. Keep all XML tags: <thoughts>, <plan>, <title>, <summary>, <scratch_pad>, <response>, etc.

Ensure you strictly follow the rules.
###======###
\end{lstlisting}

\subsection{An Example of Thai Thinking Trace Generated By Typhoon T1}\label{appendix:thai_thinking}

\begin{figure}
    \centering
    \includegraphics[width=0.6\linewidth]{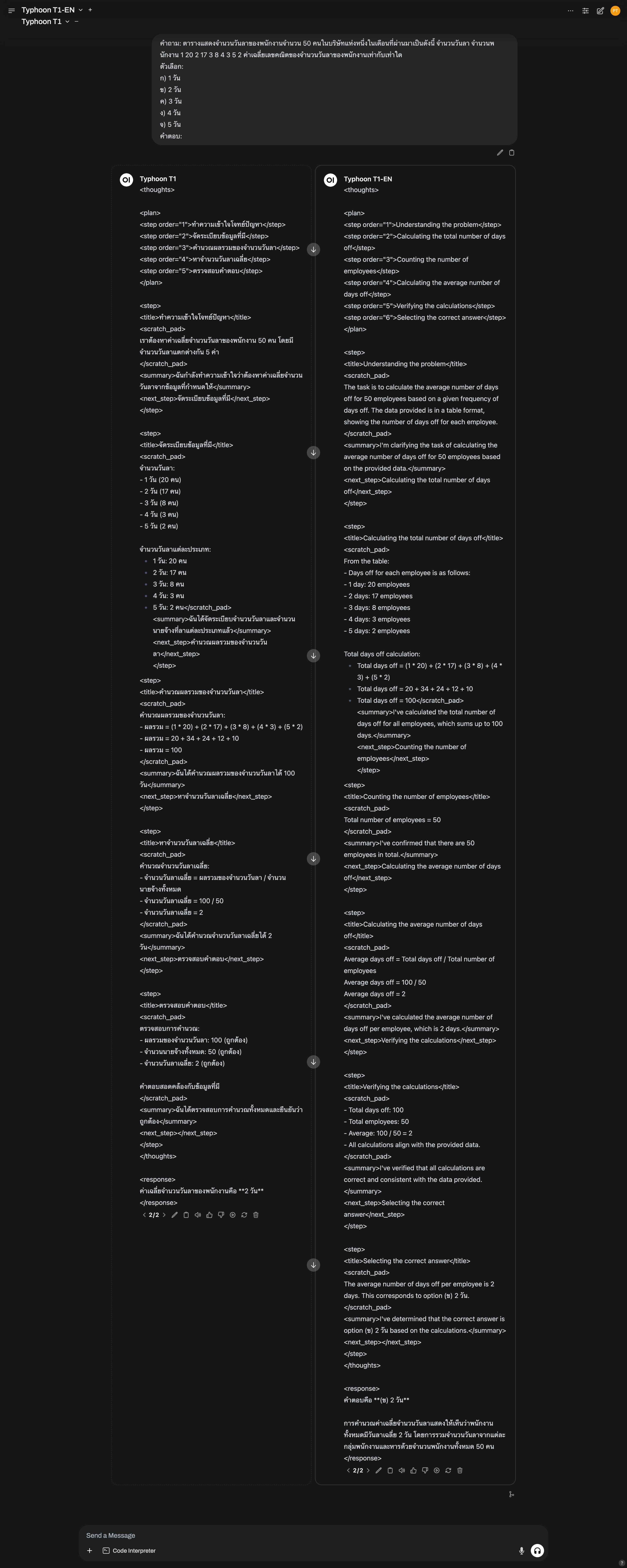}
    \caption{This figure shows Typhoon T1's Thai thinking trace and Typhoon T1-EN's English thinking trace.}
    \label{fig:t1_reasoning}
\end{figure}

We prompted Typhoon T1-EN and Typhoon T1 using the following prompt at temperature 0: \url{https://pastes.io/thai-prompt} (basic mathematics problem in Thai), and obtained the following results, also shown in \Cref{fig:t1_reasoning}:

\begin{itemize}
    \item \textbf{Typhoon T1-EN}: \url{https://pastes.io/english-thinking}
    \item \textbf{Typhoon T1}: \url{https://pastes.io/thai-thinking}
\end{itemize}

We can now observe that the thinking trace—which previously required Typhoon T1-EN to think in English—is now in Thai, and both results are correct. We use the following functions to count number of English and Thai characters for \Cref{sec:thai_thinking}.

\begin{lstlisting}[language=Python,frame=single]
def count_thai_char(text):
    count = 0
    for char in text:
        if ord(char) >= 3584 and ord(char) <= 3711:
            count += 1
    return count

def count_eng_char(text):
    count = 0
    for char in text:
        if ord(char) >= 65 and ord(char) <= 122:
            count += 1
    return count
\end{lstlisting}

\subsection{Force Thinking in Language Prompts}\label{appendix:force_prompt}

We perform an experiment to force a reasoning model to think in a specific language in \Cref{sec:force_thinking}. The following prompt is used to elicit such model behavior. Note that \lstinline|{lang}| is replaced with ``Thai'' or ``English''.

\begin{lstlisting}[frame=single]
You always think in {lang} for the <thoughts>. <thoughts> only contains {lang} language. However, the final <response> should be in the same language as the following query.
\end{lstlisting}

\end{document}